\documentclass[preprint,12pt]{elsarticle}

\usepackage{amsmath}

\usepackage{amssymb}
\usepackage{amsthm}

\journal{Pattern Recognition}

\newtheorem{thm}{Theorem}
\newtheorem{cor}[thm]{Corollary}
\newtheorem{lem}[thm]{Lemma}

\newcommand{\probover}[2]{Pr_{#1}\left\{#2\right\}}
\newcommand{\probf}[1]{\probover{F \sim D^n}{#1}}

\newcommand{\probplain}[1]{\probover{}{#1}}
\newcommand{\meanover}[2]{E_{#1}\left\{#2\right\}}

\newcommand{\mean}[1]{E \left\{#1\right\}}

\newcommand{\sset}[1]{\{#1\}}

\newcommand{\gc}{g} 

\newcommand{\be}{\begin{equation}}
\newcommand{\ee}{\end{equation}}
\newcommand{\kth}{k^{th}}
\newcommand{\ith}{i^{th}}

\newcommand{\bigo}[1]{\hbox{O} \left( #1 \right)}

\newcommand{\inn}{I(b_{i-1})}
\newcommand{\fis}{f_{i,\sigma}}

\newcommand{\phone}{\hat{p}_{I}}

\newcommand{\gw}[1]{g_{-\sset{#1}}}
\newcommand{\gws}[1]{g_{-{#1}}}
\newcommand{\wrong}[1]{\overline{#1}}
\newcommand{\wrongs}{\wrong{\gws{S}}}
\newcommand{\wrongsi}{\wrong{\gws{(S \cup \sset{i})}}}

\begin{document}

\begin{frontmatter}


\title{Speculate-Correct Error Bounds for $k$-Nearest Neighbor Classifiers}


\author[ebx]{Eric Bax}
\author[lw]{Lingjie Weng}
\author[xt]{Xu Tian}

\address[ebx]{Verizon (baxhome@yahoo.com)}
\address[lw]{LinkedIn (lingjieweng@gmail.com)}
\address[xt]{Sorin Capital Management (tianxu03@gmail.com)}

\begin{abstract}
We introduce the speculate-correct method to derive error bounds for local classifiers. Using it, we show that $k$-nearest neighbor classifiers, in spite of their famously fractured decision boundaries, have exponential error bounds with $\bigo{\sqrt{(k  + \ln n)/n}}$ error bound range for $n$ in-sample examples.
\end{abstract}


\end{frontmatter}


\section{Introduction}
Local classifiers use only a small subset of their examples to classify each input. The best-known local classifier is the nearest neighbor classifier. To classify an example, a $k$-nearest neighbor ($k$-nn) classifier uses a majority vote over the $k$ in-sample examples closest to the example. We assume $k$ is odd, and we assume binary classification. For general information on $k$-nn classifiers, see the books by Devroe et al. \cite{devroye96}, Duda et al. \cite{duda01}, and Hastie et al. \cite{hastie09}. Deriving error bounds for $k$-nn classifiers is a challenge, because they can have extremely fractured decision boundaries, making approaches based on hypothesis class size ineffective. 

The error bounds in this paper are probably approximately correct (PAC) bounds, consisting of a range of error rates and a bound on the probability that the actual out-of-sample error rate is outside the range. An effective PAC bound has a small range and a small bound failure probability. PAC error bounds include bounds based on Vapnik-Chervonenkis (VC) dimension \cite{vapnik71}, bounds for concept learning by Valiant \cite{valiant84}, compression-based bounds by Littlestone and Warmuth \cite{littlestone86}, Floyd and Warmth \cite{floyd95}, Blum and Langford \cite{blum03}, and Bax \cite{bax_compression}, and bounds based on worst likely assignments \cite{baxcallejas}. Langford \cite{langford05} gives an overview and comparison of some types of PAC bounds.  


A previous method by Devroe and Wagner \cite{devroye79} (also presented in Devroe et al. \cite{devroye96} p. 415, Theorem 24.5) produces a $k$-nn exponential error bound with range $\bigo{(k/n)^{1/3}}$. Another method \cite{bax12} has expected error bound range $\bigo{(k/n)^{2/5}}$. (Exponential error bounds have range proportional to $\ln (1/\delta)$ as bound failure probability $\delta \rightarrow 0$.)

The great conundrum of classifier validation is that we want to use data that are independent of the classifier to estimate its error rate, but we also want to use all available data for the classifier. At each step, speculate-correct assumes that this problem does not exist, at least for some of the in-sample data. In subsequent steps, it corrects for its sometimes-false earlier assumptions. As it does this, the number of corrections grows, but the size of each correction shrinks. 

To illustrate, suppose we use two withheld data sets: $V_1$ and $V_2$. Let $g$ be the full classifier; our goal is to bound its error rate: $\probplain{\wrong{g}}$. (Use $\probplain{}$ to indicate probability over out-of-sample examples, and use a bar on top to indicate classifier error.) Let $\gws{S}$ be the classifier formed by withholding the data sets indexed by $S$.  For example, $\gw{1}$ is all in-sample examples except those in $V_1$. Then the speculate-correct process is:
 
\begin{enumerate}
\item Speculate that withholding $V_1$ does not affect classification: $\forall x: g = \gw{1}$. Compute $\probover{V_1}{\wrong{\gw{1}}}$ as our initial estimate of $\probplain{\wrong{g}}$. (Use $\probover{V_i}{}$ to indicate empirical rate over examples in $V_i$ -- also called an empirical mean.) When our speculation is false ($g \not= \gw{1}$), there is bias, because then we have estimated $\probplain{\wrong{\gw{1}}}$ instead of $\probplain{\wrong{g}}$:
\be
\probplain{\wrong{\gw{1}}} = \probplain{g = \gw{1} \land \wrong{\gw{1}}} + \probplain{g \not= \gw{1} \land \wrong{\gw{1}}},
\ee
and
\be
\probplain{\wrong{g}} = \probplain{g = \gw{1} \land \wrong{g}} + \probplain{g \not= \gw{1} \land \wrong{g}}.
\ee
The first terms are the same, since $(g = \gw{1}) \implies (\wrong{g} = \wrong{\gw{1}})$, but not the second ones. To correct the bias, we need to: 
  \begin{enumerate}
  \item Subtract an estimate of $\probplain{g \not= \gw{1} \land \wrong{\gw{1}}}$.
  \item Add an estimate of $\probplain{g \not= \gw{1} \land \wrong{g}}$.
  \end{enumerate}

\item Now speculate that $\forall x: g = \gw{2}$ and $\gw{1} = \gw{1,2}$. Then use $V_2$ to correct according to (a) and (b) from Step 1:
  \begin{enumerate}
  \item Subtract $\probover{V_2}{\gw{2} \not= \gw{1,2} \land \wrong{\gw{1,2}}}$.
  \item Add $\probover{V_2}{\gw{2} \not= \gw{1,2} \land \wrong{\gw{2}}}$.
  \end{enumerate}
When the speculation for this step holds, withholding $V_2$ does not affect the corrections:
\be
\probover{V_2}{\gw{2} \not= \gw{1,2} \land \wrong{\gw{1,2}}} = \probover{V_2}{g \not= \gw{1} \land \wrong{\gw{1}}},
\ee
and 
\be
\probover{V_2}{\gw{2} \not= \gw{1,2} \land \wrong{\gw{2}}} = \probover{V_2}{g \not= \gw{1} \land \wrong{g}}.
\ee
The remaining bias terms require Step 1 and 2 speculations to fail simultaneously. For (a), the bias terms are:
\be
- \probplain{g \not= \gw{1} \land \gw{1} \not= \gw{1,2} \land \wrong{\gw{1,2}}}
\ee
\be
+ \probplain{g \not= \gw{1} \land \gw{1} \not= \gw{1,2} \land \wrong{\gw{1}}}
\ee
For (b), they are:
\be
- \probplain{g \not= \gw{1} \land g \not= \gw{2} \land \wrong{\gw{2}}}
\ee
\be
+ \probplain{g \not= \gw{1} \land g \not= \gw{2} \land \wrong{g}}
\ee
\end{enumerate}

Continuing this for $r$ steps, with $r$ withheld data sets, produces a sum of $2^r-1$ estimates. All remaining bias depends on simultaneous failure of $r$ speculations, but there are $2^r$ bias terms. For $k$-nn, speculation can only fail for Step $i$ if $V_i$ has a nearer neighbor to $x$ than its $\kth$ nearest neighbor among the in-sample examples not in any validation set. So the bias is at most $2^r$ times the probability that $x$ has a nearer neighbor in every validation set than the $\kth$ nearest neighbor among the other in-sample examples. 

To produce effective error bounds, we must use withheld data sets small enough to make the probability of $r$ simultaneous speculation failures small, and yet large enough that the sum of $2^r-1$ estimates is likely to have a small deviation from the sum that it estimates. (Using Hoeffding bounds \cite{hoeffding63}, the range for the difference between each estimate $\probover{V_i}{}$ and its corresponding out-of-sample probability $\probplain{}$ is $\bigo{\frac{1}{\sqrt{|V_i|}}}$.) We show that an appropriate choice of withheld data set size gives error bound range:
\be
\bigo{n^{-\frac{r}{2r+1}} \sqrt{k + r}},
\ee
and, for a choice of $r$ based on $n$, the range is
\be
\bigo{\sqrt{(k  + \ln n)/n}}.
\ee

The next section formally introduces the speculate-correct method to produce error bounds for local classifiers. Section \ref{results_section} applies the method to $k$-nn classifiers. Section \ref{computation_section} shows how to compute the bounds. Section \ref{discussion_section} concludes with potential directions for future work. 

\section{Speculate-Correct} \label{definitions_section}
Let $F$ be the full set of $n$ in-sample examples $(x,y)$, drawn i.i.d. from a joint input-output distribution $D$. Inputs $x$ are drawn from an arbitrary domain, and outputs $y$ are drawn from $\sset{0,1}$ (binary classification). Assume there is some ordering of the examples in $F$, so that we may refer to examples 1 to $n$ in $F$, treating $F$ as a sequence.   

Select $r>0$ and $m>0$ such that $rm \leq n-k$. For each $i \in {1, \ldots, r}$, let validation subset $V_i$ be the $i$th subset of $m$ examples in $F$. For example, if $r = 2$ and $m = 1000$, then $V_1$ is the first thousand examples in $F$ and $V_2$ is the second thousand. Let validation set $V = V_1 \cup \ldots \cup V_r$. For convenience, define $R \equiv \sset{1, \ldots, r}$. For $S \subseteq R$, let $V_S$ be the union of validation subsets indexed by $S$.


Our PAC error bounds have probability of bound failure over draws of $F$. Let the subscript $F \sim D^n$ denote a probability or expectation over draws of $F$. We use no subscript for probabilities or expectations over out-of-sample examples $(x,y) \sim D$. For example, 
\be
p^* \equiv \probplain{\wrong{\gc}}
\ee
denotes the out-of-sample error rate of $\gc$, and it is the quantity we wish to bound. (It is sometimes called the conditional error rate, because it is the error rate conditioned on a set of in-sample examples $F$ rather than the expected error rate over draws of $F$.)


Let $A_i = \sset{1, \ldots, i}$. Let $a_1, \ldots, a_r$ be any series of conditions such that
\be
a_i(x) \implies \forall S \subseteq A_{i-1}: \gws{(S \cup \sset{i})}(x) = \gws{S}(x),
\ee
i.e., $a_i(x)$ implies that for any classifier formed by withholding any subset of $\sset{V_1, \ldots, V_{i-1}}$, withholding $V_i$ too does not alter the classification of $x$. 

Let $b_i = \neg a_1 \land \ldots \land \neg a_i$. Define $b_0$ to be true. The following theorem generalizes the speculate-correct formula for $r=2$ that we developed in the previous section.  

\begin{thm} \label{general_bound} 
\be
\forall r \geq 0: p^* = \sum_{i=1}^{r} \sum_{S \subseteq A_{i-1}} (-1)^{|S|} \probplain{b_{i-1} \land \wrongsi} + \sum_{S \subseteq R} (-1)^{|S|} \probplain{b_{r} \land \wrongs}. \label{gensmile}
\ee
\end{thm}

\begin{proof}
Use induction. The base case is $r=0$:
\be
\sum_{S \subseteq \emptyset} (-1)^{|S|} \probplain{b_{0} \land \wrongs} = \probplain{b_{0} \land \wrong{\gws{\emptyset}}} = \probplain{\wrong{g}} = p^*.
\ee

Next, to show that the result for $r$:
\be
p^* = \sum_{i=1}^{r} \sum_{S \subseteq A_{i-1}} (-1)^{|S|} \probplain{b_{i-1} \land \wrongsi} + \sum_{S \subseteq R} (-1)^{|S|} \probplain{b_{r} \land \wrongs},
\ee
implies the result for $r+1$:
\be
p^* = \sum_{i=1}^{r+1} \sum_{S \subseteq A_{i-1}} (-1)^{|S|} \probplain{b_{i-1} \land \wrongsi} + \sum_{S \subseteq A_{r+1}} (-1)^{|S|} \probplain{b_{r+1} \land \wrongs},
\ee
subtract the result for $r$ from the result for $r+1$. The difference is
\be
\sum_{S \subseteq A_{r}} (-1)^{|S|} \probplain{b_{r} \land \wrong{\gws{(S \cup \sset{r+1})}}} + \sum_{S \subseteq A_{r+1}} (-1)^{|S|} \probplain{b_{r+1} \land \wrongs} 
\ee
\be
- \sum_{S \subseteq R} (-1)^{|S|} \probplain{b_{r} \land \wrongs}.
\ee
We will show that this difference is zero. 

Since $A_{r} = R$, the first and third sums are over the same indices, so combine them:
\be
= \sum_{S \subseteq A_{r}} (-1)^{|S|} \left(\probplain{b_{r} \land \wrong{\gws{(S \cup \sset{r+1})}}} - \probplain{b_{r} \land \wrongs}\right)
\ee
\be
+ \sum_{S \subseteq A_{r+1}} (-1)^{|S|} \probplain{b_{r+1} \land \wrongs}.
\ee
Expand the first sum's probabilities around $a_{r+1}$ values:
\be
\probplain{b_{r} \land \wrong{\gws{(S \cup \sset{r+1})}}} - \probplain{b_{r} \land \wrongs}
\ee
\be
= \probplain{b_{r} \land a_{r+1} \land \wrong{\gws{(S \cup \sset{r+1})}}} + \probplain{b_{r} \land \neg a_{r+1} \land\wrong{\gws{(S \cup \sset{r+1})}}} 
\ee
\be
- \probplain{b_{r} \land a_{r+1} \land \wrongs} - \probplain{b_{r} \land \neg a_{r+1} \land \wrongs}.
\ee
The first and third terms cancel, because $a_{r+1} \implies \gws{(S \cup \sset{r+1})} = \gws{S}$. The other terms have $b_{r+1}$, since $b_{r} \land \neg a_{r+1} = b_{r+1}$. So the difference is:
\be
= \sum_{S \subseteq A_{r}} (-1)^{|S|} \left(\probplain{b_{r+1} \land \wrong{\gws{(S \cup \sset{r+1})}}} - \probplain{b_{r+1} \land \wrongs}\right)
\ee
\be
+ \sum_{S \subseteq A_{r+1}} (-1)^{|S|} \probplain{b_{r+1} \land \wrongs}.
\ee
The first sum cancels the second: for each $S$ in the first sum, the first term cancels the term for $S \cup \sset{r+1}$ in the second sum, and the second term cancels the term for $S$ in the second sum. 
\end{proof}

The formulation of the error rate in Theorem \ref{general_bound} is useful because the examples in each validation subset $V_i$ are independent of the conditions in term $i$ in the first sum. So the rates of the conditions over the validation subsets are unbiased estimates of the probabilities of those conditions over out-of-sample examples. There are no such validation data for the second sum. Instead of estimating the second sum, our error bounds bound each of its terms by $\probplain{b_r}$. We select validation subset sizes to mediate a tradeoff: large validation subsets give tight bounds on terms in the first sum, but small validation subsets make $\probplain{b_r}$ small.

\section{Error Bounds for $k$-NN Classifiers} \label{results_section}
Before introducing $k$-nn error bounds, we need a brief aside about tie-breaking. Break ties using the method from Devroe and Wagner \cite{devroye79}: assign each example $i$ in $F$ a real value $Z_i$ drawn uniformly at random from $[0,1]$ and do the same for each other draw $x$ from the input space to give it a value $Z$. If the distance from example $i$ in $F$ to an $x$ is the same as the distance from example $j$ in $F$ to $x$, then declare $i$ to be the closer example if $|Z_i - Z| < |Z_j - Z|$ or if $|Z_i - Z| = |Z_j - Z|$ and $i<j$. Otherwise declare example $j$ to be the closer example. This method returns the same ranking of distances to examples in $F$ for the same input $x$ every time the distances are measured, and it uses position within $F$ to break a tie with probability zero.  

Apply the speculate-correct concept to $k$-nn:

\begin{cor} \label{gathers}
Let $a_i(x)$ be the condition that $V_i$ does not have an example closer to $x$ than the $kth$ nearest neighbor to $x$ in $F-V$. Let
\be
\forall 1 \leq i \leq r: f_i(x,y) = \inn \sum_{S \subseteq A_{i-1}} (-1)^{|S|} I\left( \wrongsi \right)
\ee
and 
\be
f_{r+1}(x,y) = \inn \sum_{S \subseteq A_{r}} (-1)^{|S|} I\left( \wrongs \right),
\ee
where $I()$ is the indicator function: one if its argument is true and zero otherwise. Then
\be
p^* = \sum_{i=1}^{r+1} \mean{f_i}. \label{hiho}
\ee
\end{cor}

\begin{proof}
Our $a_i$ for $k$-nn meet the conditions of Theorem \ref{general_bound}.
\end{proof}

Next, we show that $p^*$ is the average of the RHS of Equation \ref{hiho} from Corollary \ref{gathers} over all permutations of the in-sample examples. Permuting the examples places different examples into the validation subsets because the $\ith$ validation subset is the $\ith$ $m$ examples. We will use permutations to ensure that $b_1, \ldots, b_r$ are rare enough to provide small error bound ranges. 

Without permutations, even $b_r$ may not be rare. For example, in-sample examples $m, 2m, \ldots, rm$ may all be close to much of the input distribution, and the other in-sample examples may be far. Without permutations, we can develop a bound, but we can only show that it has a small error bound range in expectation. Permutations guarantee that the expectation is realized. In the next section, we show how to compute permutation-based bounds efficiently. 

\begin{lem} \label{perms}
Let $P$ be the set of permutations of $1, \ldots, n$. For each $\sigma \in P$, let $\sigma F$ be $F$ permuted according to $\sigma$: example $j$ of $\sigma F$ is the example of $F$ indexed by element $j$ of $\sigma$. Let $f_{i,\sigma}$ be $f_i$, but with $F$ replaced by $\sigma F$, so that for $i \in R$, $V_i$ consists of the $\ith$ $m$ examples in $\sigma F$. Then

\be
p^* = \meanover{\sigma \in P}{\sum_{i=1}^{r+1} \mean{f_{i,\sigma}}}. \label{lpseq}
\ee
\end{lem}

\begin{proof}
Corollary \ref{gathers} holds for each partition of $F$ into $r$ size-$m$ subsets $V_1, \ldots, V_r$ and $F-V$. Each permutation of $F$ uses one of these partitions to define $f_{i,\sigma}$. So the outer expectation is over quantities that are each $p^*$. 
\end{proof}

Separate the RHS of Equation \ref{lpseq} into terms with $i \in R$ and a term with $i=r+1$:
\be
p^* = \left(\sum_{i=1}^r \meanover{\sigma \in P}{\mean{\fis}}\right)+ \meanover{\sigma \in P}{\mean{f_{r+1,\sigma}}}
\ee
\be
= p_I + p_{II}.
\ee
To bound $p_{II}$, notice that 
\be
\forall (x,y): |f_{r+1,\sigma}| \leq 2^r I(b_r| \sigma),
\ee
using $\cdot|\sigma$ to denote a condition or set with $\sigma F$ in place of $F$. So, since
\be
p_{II} = \meanover{\sigma \in P}{\mean{f_{r+1,\sigma}}}
\ee
and
\be
f_{r+1,\sigma} = I\left(b_r\right) \sum_{S \subseteq A_r} (-1)^{|S|} I\left( \wrongs \right),
\ee
\be
|p_{II}| \leq 2^r \meanover{\sigma \in P}{\mean{I(b_r|\sigma)}}.
\ee
Exchange the order of expectations:
\be
|p_{II}| \leq 2^r \mean{\meanover{\sigma \in P}{I(b_r|\sigma)}}. \label{ptwo}
\ee
For each $(x,y)$ and $F$, the inner expectation is the same with probability one. It is the probability that a random permutation places at least one example in each of $V_1, \ldots, V_r$ that is closer to $(x,y)$ than the $\kth$ closest example to $(x,y)$ in $F-V$. Since determining positions in a permutation is equivalent to drawing positions at random without replacement, it is the same as the probability of drawing at least one element from each set $\sset{1, \ldots, m}, \ldots \sset{(r-1)m + 1, \ldots, rm}$ before drawing $k$ elements from $\sset{rm+1, \ldots, n}$, when drawing uniformly at random without replacement from $\sset{1, \ldots, n}$. 

The probability of drawing $k$ elements from $\sset{rm+1, \ldots, n}$ before drawing any from one specific set in $\sset{1, \ldots, m}, \ldots \sset{(r-1)m + 1, \ldots, rm}$ is
\be
\left(\frac{n-rm}{n-rm+m}\right) \left(\frac{n-rm-1}{n-rm+m-1}\right) \cdots \left(\frac{n-rm-(k-1)}{n - rm +m - (k-1)}\right).
\ee
Similarly, the probability of drawing $k$ elements from $\sset{rm+1, \ldots, n}$ before drawing any elements from any of $i$ specific sets in $\sset{1, \ldots, m}, \ldots \sset{(r-1)m + 1, \ldots, rm}$ is
\be
\left(\frac{n-rm}{n-rm+im}\right) \left(\frac{n-rm-1}{n-rm+im-1}\right) \cdots \left(\frac{n-rm-(k-1)}{n - rm +im - (k-1)}\right).
\ee
So, by inclusion and exclusion, the probability of drawing at least one element from every set in $\sset{1, \ldots, m}, \ldots \sset{(r-1)m + 1, \ldots, rm}$ before drawing $k$ examples from $\sset{rm+1, \ldots, n}$ is:
\be
\probover{\sigma \in P}{b_r|\sigma} = \sum_{i=0}^r (-1)^i {{r}\choose{i}} \prod_{j=0}^{k-1} \left(\frac{n-rm-j}{n-rm+im-j}\right). \label{crx_exact}
\ee

The following lemma bounds this probability:
\begin{lem} \label{donut_hole}
\be
\forall x: \probover{\sigma \in P}{b_r|\sigma} \leq \left( \frac{e(k+r-1)m}{n} \right)^r.
\ee
\end{lem}

\begin{proof}
Define $d_R(x)|\sigma$ to be the condition that the $k+r-1$ nearest neighbors to $x$ in $F$ include at least $r$ examples from the validation set $V|\sigma$. Condition $d_R|\sigma$ is a necessary condition for $b_r|\sigma$.  
\be
\forall x: \probover{\sigma \in P}{d_R|\sigma} \geq \probover{\sigma \in P}{b_r|\sigma}.
\ee
The probability of $d_R|\sigma$ over $\sigma \in P$ is the same as the probability of drawing $k+r-1$ samples from $1, \ldots, n$ uniformly without replacement and having at least $r$ of those samples have values $rm$ or less. (The samples are the indices in $\sigma F$ of the $k+r-1$ nearest neighbors to $x$.) So the probability of $d_R$ is the tail of a hypergeometric distribution:
\be
\forall x: \probover{\sigma \in P}{d_R|\sigma} = \sum_{i=r}^{k+r-1} \frac{{{k+r-1}\choose{i}} {{n-(k+r-1)}\choose{rm-i}}}{{{n}\choose{rm}}}. 
\ee
Using a hypergeometric tail bound from Chv\'atal \cite{chvatal79} (see also Skala \cite{skala13}), this is
\be
\leq \left( \frac{(k+r-1)m}{n} \right)^r \left[ \left(1 + \frac{1}{m-1} \right) \left(1 - \frac{k+r-1}{n} \right) \right]^{(m-1)r}
\ee
\be
\leq \left( \frac{(k+r-1)m}{n} \right)^r \left[ \left(1 + \frac{1}{m-1} \right)^{m-1} \right]^r
\ee
\be
\leq \left( \frac{(k+r-1)m}{n} \right)^r e^r.
\ee
\end{proof}

Let
\be
\epsilon_{II} = \left( \frac{2e(k+r-1)m}{n} \right)^r \label{etwo_def}.
\ee
Then, using Lemma \ref{donut_hole} and Inequality \ref{ptwo}, 
\be
|p_{II}| \leq \epsilon_{II}. \label{ptwo_bound}
\ee

Now consider $p_I$:
\be
p_I = \sum_{i=1}^r \meanover{\sigma \in P}{\mean{\fis}}. \label{pone_sum}
\ee
For each $\sigma \in P$ and $i \in R$, the examples in $V_i|\sigma$ are independent of the function $\fis$, so we can use empirical means over $(x,y) \in V_i|\sigma$ to estimate means over $(x,y) \sim D$. The following two lemmas give a bound for $p_I$ based on this approach.

\begin{lem} \label{poneq}
$\forall i \in R$, let $|V_i| = m$. Let $M$ be the set of size-$m$ subsets of $F$: $M = \sset{Q | Q \subseteq F \land |Q|=m}$. Let $P(Q,i)$ be the set of permutations of $1, \ldots, n$ that have set $Q$ as validation set $V_i$ in $\sigma F$: $P(Q,i) = \sset{\sigma | (V_i | \sigma)=Q}$. Then
\be
p_I = \meanover{Q \in M}{\sum_{i=1}^r \meanover{\sigma \in P(Q,i)}{\mean{\fis}}} \label{pone_qm}
\ee
\end{lem}

\begin{proof}
Compare Equations \ref{pone_sum} and \ref{pone_qm}. Equation \ref{pone_sum} averages over permutations in $P$ and $i \in R$. In Equation \ref{pone_qm}, the expectation over $Q \in M$, $i \in R$, and $P(Q,i)$ covers all permutations $P$ and $i \in R$, each with equal frequency.
\end{proof}

\begin{lem} \label{pone}
Let 
\be
\hat{p}_Q = \meanover{(x,y) \in Q}{\sum_{i=1}^r \meanover{\sigma \in P(Q,i)}{\fis}}, \label{phq_def}
\ee
and let
\be
\phone = \meanover{Q \in M}{\hat{p}_Q}. \label{phone_def}
\ee
Then
\be
\forall \delta>0: \probover{F \sim D^n}{|p_I - \phone| \geq \epsilon_I} \leq \delta, \label{pone_bound}
\ee
where
\be
\epsilon_I \leq \frac{1}{\sqrt{m}} \left(\frac{1}{1 - \frac{2e(k+r-2)m}{n}}\right) \frac{1}{\sqrt{2}} \left(1.06 \sqrt{\ln \frac{2}{\delta}} + 3.22\right). \label{epsone}
\ee
\end{lem}

\begin{proof}
Let 
\be
p_Q = \mean{\sum_{i=1}^r \meanover{\sigma \in P(Q,i)}{\fis}}.
\ee
Then
\be
p_I = \meanover{Q \in M}{p _Q},
\ee
since this is Inequality \ref{pone_qm} from Lemma \ref{poneq}, with a different order of expectations. 

For each $Q \in M$, we will use each $\hat{p}_Q$ to bound each $p_Q$, using the fact the examples in $Q$ are independent of $p_Q$. First, we need to bound the range of terms in the expectations $p_Q$ and $\hat{p}_Q$:
\be
\sum_{i=1}^r \meanover{\sigma \in P(Q,i)}{\fis}.
\ee

Recall that 
\be
\fis = I(b_{i-1}|\sigma) \sum_{S \subseteq A_{i-1}} (-1)^{|S|} I\left(\wrong{\gws{S \cup \sset{i}}} | \sigma\right),
\ee
So for each $(x,y)$ and $i \in R$, in the expectation
\be
\meanover{\sigma \in P(Q,i)}{\fis},
\ee
the fraction of $\sigma \in P(Q,i)$ for which $b_{i-1}|\sigma$ is the same as the probability of drawing at least one element from each of the first $m$, second $m$, ..., $(i-1)^{st}$ $m$ before drawing $k$ elements from the last $n-rm$ when drawing a sequence uniformly at random without replacement from $\sset{1, \ldots, (i-1)m, rm+1, \ldots, n}$. This is less than the probability of drawing $i-1$ elements from $\sset{1, \ldots, (i-1)m}$ before drawing $k$ elements from $\sset{rm+1, \ldots, n}$. Using the proof of Lemma \ref{donut_hole}, with $i-1$ in place of $r$, this probability is at most
\be
\left(\frac{e(k+i-2)m}{n}\right)^{i-1}.
\ee
So at most this fraction of $\sigma \in P(Q,i)$ produce nonzero $\fis$.

If $b_{i-1}|\sigma$, then $\fis$ sums over $2^{i-1}$ terms, and half are zero or one and half are zero or negative one, so
\be
\fis \in [-2^{i-2}, 2^{i-2}].
\ee
Multiplying the fraction of nonzero terms by their range bounds:
\be
\meanover{\sigma \in P(Q,i)}{\fis} \in \left[-\frac{1}{2} \left(\frac{2e(k+i-2)m}{n}\right)^{i-1}, \frac{1}{2} \left(\frac{2e(k+i-2)m}{n}\right)^{i-1}\right].
\ee
Now sum the range upper bound over $i \in R$:
\be
\sum_{i=1}^r \frac{1}{2} \left(\frac{2e(k+i-2)m}{n}\right)^{i-1}
\ee
Since $k+i-2 \leq k+r-2$, this is
\be
\leq \frac{1}{2} \sum_{i=1}^{r} \left(\frac{2e(k+r-2)m}{n}\right)^{i-1}.
\ee
Use the identity $1 + x + x^2 + \ldots + x^{r-1} = \frac{1 - x^r}{1 - x}$, with $x = \frac{2e(k+r-2)m}{n}$ (with $x$ as a number, not the input of an example as in the rest of this paper):
\be
\leq \frac{1}{2} \left(\frac{1}{1 - \frac{2e(k+r-2)m}{n}}\right).
\ee
So
\be
\sum_{i=1}^r \meanover{\sigma \in P(Q,i)}{\fis} \in \left[- \frac{1}{2} \left(\frac{1}{1 - \frac{2e(k+r-2)m}{n}}\right), \frac{1}{2} \left(\frac{1}{1 - \frac{2e(k+r-2)m}{n}}\right)\right].
\ee
Based on this range, we can apply Hoeffding bounds \citep{hoeffding63}:
\be
\forall Q \in M, \delta>0: \probover{F \sim D^n}{|p_Q - \hat{p}_Q| \geq \left(\frac{1}{1 - \frac{2e(k+r-2)m}{n}}\right) \sqrt{\frac{\ln \frac{2}{\delta}}{2m}}} \leq \delta. \label{sad_face}
\ee

Since
\be
|p_I - \phone| = |\meanover{Q \in M}{p_Q} - \meanover{Q \in M}{\hat{p}_Q}|, \label{avgm}
\ee
to bound $p_I$ using $\phone$, we need to bound the difference between the average of $p_Q$ values and the average of corresponding empirical means  $\hat{p}_Q$. According to \cite{bax_average} (Expression 12 page 8), the bound range for the difference in averages is not much larger than the average of the Hoeffding bound ranges for the individual differences over $Q$. Applying that result:
\be
\forall c>0, \delta>0: \probover{F \sim D^n}{|p_I - \phone| \geq \epsilon} \leq \delta,
\ee
where 
\be
\epsilon \leq \frac{1}{\sqrt{2 m}} \left(\frac{1}{1 - \frac{2e(k+r-2)m}{n}}\right) \left[ \sqrt{\ln \frac{2}{\delta}} \left(\frac{e^c}{e^c - 1}\right) + \sqrt{c+1} \left(\frac{e^c}{e^c - 1}\right)^2 + 1\right] \label{eps1}
\ee
(We have $\frac{2}{\delta}$ in place of the $\frac{1}{\delta}$ in \cite{bax_average}, because we have two-sided bounds.)  Set $c = 3$ to prove the lemma.
\end{proof}

Combine this bound on $p_I$ with the bound on $|p_{II}|$ to bound $p^*$:

\begin{lem} \label{combined_bound}
\be
\forall \delta>0: \probover{F \sim D^n}{|p^* - \phone| \geq \epsilon_I + \epsilon_{II}} \leq \delta. \label{base_bound}
\ee
\end{lem}

\begin{proof}
Since $p^* = p_I + p_{II}$, 
\be
|p^* - \phone| = |(p_I + p_{II}) - \phone| \leq |p_I - \phone| + |p_{II}|.
\ee
By Inequality \ref{ptwo_bound}, $|p_{II}| \leq \epsilon_{II}$. From the definition of $\epsilon_I$ (Lemma \ref{pone}),
\be
\forall \delta>0: \probover{F \sim D^n}{|p_I - \phone| \geq \epsilon_I} \leq \delta.
\ee
So
\be
\forall \delta>0: \probover{F \sim D^n}{|p_I - \phone| + |p_{II}| \geq \epsilon_I + \epsilon_{II}} \leq \delta.
\ee
\end{proof}

The following Theorem and Corollary are the main results for $k$-nn classifiers. The theorem allows $r$, $k$, and $\delta$ to depend on the number of in-sample examples, $n$. The corollary uses the bound from the theorem with an appropriate growth rate for $r$ as $n$ increases ($r \in \hbox{O}(\ln n)$). 
\begin{thm} \label{rbound}
\be
\forall \delta > 0: \probover{F \sim D^n}{|p^* - \phone| \leq \epsilon_{r}} \leq \delta,
\ee
with 
\be
\epsilon_{r} \in \hbox{O}\left( n^{-\frac{r}{2r+1}} \sqrt{(k+r)} \right).
\ee
\end{thm}

\begin{proof}
Select validation set sizes $m$ to balance $\epsilon_I$ and $\epsilon_{II}$:
\be
m = \lceil\frac{n^\frac{r}{r+\frac{1}{2}}}{2e(k+r-1)}\rceil.
\ee
Then
\be
\epsilon_{II} = \left( \frac{2e(k+r-1)m}{n} \right)^r \in \hbox{O}\left(n^{-\frac{r}{2r+1}}\right),
\ee
and
\be
\epsilon_I \leq n^{-\frac{r}{2r+1}} \left(\frac{1}{1 - n^{-\frac{1}{2r+1}}}\right) \sqrt{2e(k+r-1)} \frac{1}{\sqrt{2}} \left(1.06 \sqrt{\ln \frac{2}{\delta}} + 3.22\right).
\ee

If we allow for the possibility of $k$ and $r$ growing with $n$, then
\be
\epsilon_r = \epsilon_I + \epsilon_{II} \in \hbox{O}\left( n^{-\frac{r}{2r+1}} \sqrt{k+r} \right).
\ee
\end{proof}

\begin{cor} \label{bound}
For a choice of $r$ based on $n$, 
\be
\forall \delta > 0: \probover{F \sim D^n}{|p^* - \phone| \leq \epsilon_{*}} \leq \delta,
\ee
with 
\be
\epsilon_{*} \in \hbox{O}\left(\sqrt{(k + \ln n)/n} \right).
\ee
\end{cor}

\begin{proof}
If we set $r = \lceil \frac{1}{4}(\ln n - 2) \rceil$, then
\be
n^{-\frac{r}{2r+1}} = n^{\frac{1}{4r+2}} n^{-\frac{1}{2}} \leq n^{\frac{1}{\ln n}} n^{-\frac{1}{2}} = e n^{-\frac{1}{2}}.
\ee
So
\be
\epsilon_{*} = \epsilon_I + \epsilon_{II} \in \bigo{n^{-\frac{1}{2}} \sqrt{(k + \ln n)}}.
\ee
\end{proof}

An alternative proof of Corollary \ref{bound} uses a different value for $m$:
\begin{proof}[Alternative Proof of Corollary \ref{bound}]
Let
\be
m = \lfloor \frac{n}{2e^2(k+r-1)}\rfloor. \label{altm}
\ee
Then
\be
\epsilon_{II} = \left(\frac{2e(k+r-1)m}{n}\right)^r \leq \frac{1}{e^r}.
\ee
For $\epsilon_I$, note that
\be
m \geq \frac{n}{2e^2(k+r-1)} - 1 = \frac{n - 2e^2(k+r-1)}{2e^2(k+r-1)}.
\ee
Substitute the RHS for $m$ in Inequality \ref{epsone}:
\be
\epsilon_I \leq \left(\frac{1}{1 - \frac{1}{e}} \right) \frac{\sqrt{2e^2(k+r-1)}}{\sqrt{n - 2e^2(k+r-1)}} \frac{1}{\sqrt{2}} \left(1.06 \sqrt{\ln \frac{2}{\delta}} + 3.22\right). \label{alte1}
\ee
Let $r = \lceil \ln \sqrt{n} \rceil$. Then
\be
\epsilon_{II} \leq \frac{1}{e^{\ln \sqrt{n}}} = \frac{1}{\sqrt{n}}, \label{alte2}
\ee
and
\be
\epsilon_{I} \in \hbox{O}\left(\sqrt{(k + \ln n)/n} \right).
\ee
\end{proof}

\section{Computation} \label{computation_section}
It would be infeasible to compute the error bounds developed in this paper directly from their definitions. Instead, we can sample the bound terms to produce a bound. In this section, we outline a sampling procedure that requires O($n (\ln n)^2$) computation (in addition to identifying up to $k+r-1$ nearest neighbors in $F$ for each example in $F$) and produces a bound with range $\bigo{\sqrt{(k + \ln n)/n}}$.

Note that
\be
\phone = \meanover{\sigma \in P}{ r \meanover{i \in R}{ \meanover{(x,y) \in V_i|\sigma}{ \fis}}}.
\ee
(For reference, $\phone$ is defined in Equations \ref{phq_def} and \ref{phone_def} of Lemma \ref{pone}.) Let $P((x,y), i) = \sset{\sigma| (x,y) \in (V_i | \sigma))}$. Reordering expectations,
\be
\phone = \meanover{(x,y) \in F}{ r \meanover{i \in R}{ \meanover{\sigma \in P((x,y), i)}{ \fis}}}. \label{phoneperm}
\ee
Rewrite $\fis$ as the expectation of its terms:
\be
\fis = I(b_{i-1}) 2^{i-1} \meanover{S \subseteq A_{i-1}}{(-1)^{|S|} I(\wrongsi |\sigma)}. \label{phonefis}
\ee

Estimate $\phone$ as defined in the previous two equations by taking an empirical mean over $s$ random samples:
\be
\left((x,y), i, \sigma, S \right),
\ee
with $(x,y)$ drawn uniformly at random from $F$, $i$ uniformly at random from $R$, $\sigma$ uniformly at random from $P((x,y),i)$, and $S$ uniformly at random from the power set of $A_{i-1}$. Each sample value is
\be
r I(b_{i-1}) 2^{i-1} (-1)^{|S|} I( \wrongsi |\sigma).
\ee
Let $p'_I$ be the empirical mean of these samples.

To bound the difference between $\phone$ and its estimate $p'_I$, we will use a two-sided version of an Inequality from Maurer and Pontil \cite{maurer09}, that is derived from Bennett's Inequality \citep{bennett62}:
\be
\forall \delta>0: \probf{|\phone - p'_I| \geq \sqrt{\frac{2 v \ln \frac{2}{\delta}}{s}} + \frac{r 2^r \ln \frac{2}{\delta}}{3 s}} \leq \delta,
\ee
where $v$ is the variance of samples for the empirical mean. (The term $r 2^r$ accounts for the range of our samples: $[-r2^{i-1}, r2^{i-1}]$.) We use this inequality instead of Hoeffding's Inequality in order to take advantage of the small variance of our samples relative to their range. 

To bound $v$, note that $b_{i-1}$ has probability at most
\be
\left( \frac{e(k+i-2)m}{n}\right)^{i-1}.
\ee
So each sample has at most that probability of being nonzero. Sample range is $[-r2^{i-1}, r2^{i-1}]$. Since variance is at most the expectation of the square:
\be
v \leq \left( \frac{e(k+i-2)m}{n}\right)^{i-1} \left(2^{i-1}\right)^2 r^2 = \left( \frac{4e(k+i-2)m}{n}\right)^{i-1} r^2.
\ee
So set
\be
m < \frac{n}{4e(k+r-2)}
\ee
to ensure $v \leq r^2$. (The value for $m$ in the alternative proof of Corollary \ref{bound} meets this condition.) Then select a number of samples $s$ to achieve a desired balance between computation and accuracy. 

Computing values for samples in $p'_I$ need not involve drawing complete permutations $\sigma$. Instead, randomly determine set membership in $F-V|\sigma$, $V_i|\sigma$, ..., or $V_r|\sigma$ for neighbors of the sample $(x,y)$ and tabulate votes to determine $I(\wrongsi |\sigma)$, proceeding one neighbor at a time until the $kth$ neighbor from 
$F-V|\sigma$ is identified, as follows.

Let $N_0(x) = (x,y)$. Let $N_j(x)$ be $(x,y)$ and the $j$ nearest neighbors to $(x,y)$ in $F$. At each step, let $f = |(F-V|\sigma) \cap N_j(x)|$. For each $i \in R$, let $v_{i} = |V_i \cap N_j(x)|$. Let $b$ be the number of voting neighbors ($b$ for ``ballots") among the $j$ nearest neighbors to $(x,y)$: $b = |((F - V_i) - \cup_{h \in S} V_h |\sigma) \cap N_j(x)|$. Let $d$ be the number of those voters that have different labels than $y$.

Initially, $j=0$, $f = 0$,  $v_{i} = 1$, $\forall h \not= i: v_{h} = 0$, $b = 0$, and $d = 0$. Then, for each $j$ starting with $j=1$, select a set for the $j^{th}$ nearest neighbor at random and increment its counter: 

\be
\begin{array}{ll}
\mbox{$F-V|\sigma$ with probability $\frac{n-f - \sum_{h \in R} (m - v_{h})}{n - j}$}: & f := f+1 \\
\mbox{$V_1|\sigma$ with probability $\frac{m - v_{1}}{n - j}$}: & v_1 := v_1 +1\\
\vdots & \vdots \\
\mbox{$V_r|\sigma$ with probability $\frac{m - v_{r}}{n - j}$}: & v_r := v_r +1 \\
 \end{array}
\ee

If $b<k$ (fewer than $k$ votes cast) and the set is $F-V|\sigma$ or $V_h|\sigma$ for $h \not = i$ and $h \not\in S$, then $b := b+1$ (another ballot is cast) and if the label of the $j^{th}$ nearest neighbor is not equal to $y$, then $d:=d+1$ (another disagreeing vote). Stop when $f=k$, and return the sample value:
\be
r I(\forall h < i: v_h>0) 2^{i-1} (-1)^{|S|} I(d \geq \frac{k+1}{2}).
\ee

This method may require up to O($rm+k$) computation per sample, because it is possible (though extremely unlikely) for an example to have all validation examples in $V|\sigma$ as nearer neighbors than the $\kth$ nearest neighbor from $F-V|\sigma$. To reduce worst-case computation, stop computation for a sample if $r$ neighbors are assigned to validation sets before the $\kth$ neighbor is assigned to $F-V|\sigma$ (that is, if $f<k$ and $v_1 + \ldots + v_r = r$), and return zero as the value for the sample. 

Recall that $d_R(x)|\sigma$ (defined in Lemma \ref{donut_hole}) is the condition that there are at least $r$ nearer neighbors to $x$ in $V|\sigma$ than the $\kth$ nearest neighbor in $F-V|\sigma$. Instead of sampling $\phone$, the modified procedure samples a modified $\phone$ that sets terms to zero if they meet condition $d_R|\sigma$. (Define $\tilde{p}_I$ to be this modified $\phone$.) To account for the lost terms, apply Lemma \ref{donut_hole}. (It applies to our $x$ from $(x,y) \in V_i|\sigma$ as well as to random $(x,y) \sim D$, because having our $(x,y)$ in a validation set only decreases the probability of $d_R|\sigma$.) So
\be
\forall i \in R: \probover{\sigma \in P((x,y), i)}{d_R|\sigma} \leq \left( \frac{e(k+r-1)m}{n} \right)^r.
\ee
Since the maximum absolute value of each of these terms is $2^{i-1}$, the sum of the lost terms is at most
\be
\sum_{i=1}^{r} 2^{i-1} \left( \frac{e(k+r-1)m}{n} \right)^r < 2^r \left( \frac{e(k+r-1)m}{n} \right)^r = \epsilon_{II}.
\ee
So add another $\epsilon_{II}$ to bounds to account for the lost terms. 

The modified procedure requires only O($k + r$) computation for each sample, beyond any computation required to find the $k+r-1$ nearest neighbors from $F - \sset{(x,y)}$ to the sample's example $(x,y)$. Using $s=r n$ samples, if $r + k \in \hbox{O}(\ln n)$, then the modified sample value procedure requires O($n (\ln n)^2$) computation in addition to neighbor-finding.

Now combine sampling, the modified sample value computation procedure, and Corollary \ref{bound} to form an error bound that can be computed efficiently.  Let $s = r n$. Use $\frac{\delta}{2}$ as the bound failure probability for the bound from Corollary \ref{bound},  and use $\frac{\delta}{2}$ as the bound failure probability for the sampling procedure. (The reasoning about the variance of sample values also applies to the modified procedure, since it zeros some sample values and leaves the rest unchanged.) Use the $m$ and $r$ values from the alternative proof of Corollary \ref{bound}: Equation \ref{altm} and $r = \lceil \ln \sqrt{n} \rceil$. Let $\tilde{p}'_I$ be average of the sample values from the modified sampling method. Then we have bound:
\be
\forall \delta>0: \probover{F \sim D^n}{|p^* - \tilde{p}'_I| \geq \tilde{\epsilon}_I + 2 \epsilon_{II} + \sqrt{\frac{2 \lceil \ln \sqrt{n} \rceil \ln \frac{4}{\delta}}{n}} + \frac{2^{\lceil \ln \sqrt{n} \rceil } \ln \frac{4}{\delta}}{3 n}} \leq \delta,
\ee
where $\tilde{\epsilon}_I$ is $\epsilon_I$ from Inequality \ref{alte1} with $\frac{\delta}{2}$ substituted for $\delta$:
\be
\tilde{\epsilon}_I \leq \left(\frac{1}{1 - \frac{1}{e}} \right) \frac{\sqrt{2e^2(k+r-1)}}{\sqrt{n - 2e^2(k+r-1)}} \frac{1}{\sqrt{2}} \left(1.06 \sqrt{\ln \frac{4}{\delta}} + 3.22\right),
\ee
and $\epsilon_{II}$ is from Inequality \ref{alte2}:
\be
\epsilon_{II} \leq \frac{1}{e^{\ln \sqrt{n}}} = \frac{1}{\sqrt{n}}. 
\ee
The bound range:
\be
\tilde{\epsilon}_I + \frac{2}{\sqrt{n}} + \sqrt{\frac{2 \lceil \ln \sqrt{n} \rceil \ln \frac{4}{\delta}}{n}} + \frac{2^{\lceil \ln \sqrt{n} \rceil } \ln \frac{4}{\delta}}{3 n}
\ee
is
\be
\bigo{\sqrt{(k + \ln n)/n}}.
\ee 

\ref{dynprog} presents methods to compute rather than estimate $\phone$ or $\tilde{p}'_I$. The method to compute $\tilde{p}'_I$ requires O($n \ln n$) computation, like sampling, but it requires O($(\ln n)^4$) space, and it is more complicated than sampling.

\section{Conclusion} \label{discussion_section}
We have shown that $k$-nearest neighbor classifiers have exponential PAC error bounds with 
\be
\bigo{\sqrt{(k + \ln n)/n}}.
\ee 
error bound ranges. The bounds are quite general. They apply to any type of inputs, because they are based on probability rather than geometry.  Also, they have no terms that increase with the number of dimensions or other properties of the input space. The bounds also do not require the $k$-nn classifier's method to compute distances among examples to be symmetric or obey the triangle inequality -- it need not be a metric in the mathematical sense. It can be any function on two example inputs that returns a number. 

It may be possible to improve the error bound by using a tighter bound on the probability that multiple validation subsets have neighbors to an input that are closer than the $k$ nearest neighbors among the non-validation in-sample examples. Our bound does not require that each validation subset have such a neighbor, only that collectively they have as many such neighbors as the number of validation subsets. Equation \ref{crx_exact} enforces that condition. In practice, use it in place of Lemma \ref{donut_hole}. 

We average bounds over all choices of validation subsets so that we can prove the resulting bound has a small range. If, instead, we use a single random choice of validation subsets, then we can also produce an exponential PAC error bound. To do this, use each validation set $V_i$ to validate $f_i()$, and use a random subset of the remaining in-sample examples to validate the rate of all validation subsets having a neighbor closer to an input than the $\kth$ nearest neighbors among the other in-sample examples. (In a transductive setting \cite{vapnik98}, or if unlabeled inputs are otherwise available, use them for this validation.) This bound has $\bigo{\sqrt{(k + \ln n)/n}}$ in expectation. We average over choices of validation subsets to guarantee that we realize the expectation. 

We showed how to use sampling to ``estimate the estimates" of the error bounds. We also showed (in the appendix) an efficient, but more complex and space-consuming, method to compute an estimate. It may be possible to improve or simplify that procedure by gathering terms in a different way. 

In the future, it would be interesting to extend the $k$-nearest neighbor error bounds from this paper to cover selection of a distance metric from a parameterized set of ``hypothesis" metrics \citep{kedem12}. One approach might be to use uniform bounds of the type derived in this paper over the class of potential metrics. The bounds might depend on some notion of the complexity of the class of potential metrics. 

Finally, it would be interesting to apply the speculate-correct technique from this paper to derive error bounds for classifiers other than nearest neighbors. Other local classifiers include some collective classifiers \citep{sen2008,macskassy2007}, such as network classifiers based only on neighbors or neighbors of neighbors in a graph. (For some background on error bounds for network classifiers, refer to \cite{london12,li12,bax13}.) It may also be possible to apply the speculate-correct method to other types of classifiers that are typically based on small subsets of the in-sample examples, such as support vector machines \citep{vapnik98,cristianini00,joachims02} and set-covering machines \citep{marchand01}.
 

\appendix

\section{Method to Compute $\phone$ and $\tilde{p}'_I$.} \label{dynprog}
By gathering terms rather than sampling, we can compute $\phone$ and $\tilde{p}_I$ exactly. In this appendix, we show how to compute $\phone$ exactly and how to compute $\tilde{p}'_I$ in O($n \ln n$) time and O($(\ln n)^4$) space, assuming $k+r \in \hbox{O}(\ln n)$ and ignoring any time and space required to find the $k+r$ nearest neighbors to each in-sample example. The methods in this section are inspired by a similar approach for a single validation subset by \cite{mullin00}.

Recall from Equations \ref{phoneperm} and \ref{phonefis} that
\be
\phone = \meanover{(x,y) \in F}{ \meanover{i \in R}{ \meanover{\sigma \in P((x,y), i)}{ \fis}}},
\ee
and
\be
\fis = I(b_{i-1}) 2^{i-1} \meanover{S \subseteq A_{i-1}}{(-1)^{|S|} I(\wrongsi |\sigma)}. 
\ee
Use the symmetry of permutations over same-size subsets $S$ to compute only for $S = \sset{1, \ldots, |S|}$, and use $s$ to index values of $|S|$. Note that 
\be
\probover{S \subseteq A_{i-1}}{|S|=s} = {{i-1}\choose{s}} 2^{-(i-1)}.
\ee
Let $A_s = \sset{1, \ldots, s}$. Let
\be
p_{(x,y),i} =  \probover{\sigma \in P((x,y), i)}{b_{i-1} \land \wrong{\gws{A_s \cup \sset{i}}} |\sigma}. \label{triangle2}
\ee
Then
\be
\phone = \meanover{(x,y) \in F}{\sum_{i=1}^{r} \sum_{s=0}^{i-1}  {{i-1}\choose{s}} (-1)^s p_{(x,y),i}}. \label{triangle3}
\ee

Refer to the $j^{th}$ nearest neighbor to $(x,y)$ in $F - \sset{(x,y)}$ as neighbor $j$. Let $c_{t,u,v}(\sigma)$ be the condition that a permutation $\sigma$ assigns the neighbors to $(x,y)$ to sets $F, V_1, \ldots, V_r|\sigma$ such that there are exactly $k$ voters (in $F - (V_i \cup V_1 \ldots V_r)|\sigma$) among neighbors 1 to $t$, neighbor $t$ is a voter, there are $k$ neighbors from $F-V|\sigma$ among neighbors 1 to $u$, neighbor $u$ is from $F-V|\sigma$, and there are $v$ voters among neighbors 1 to $u$. Let 
\be
p_{t,u,v} = \probover{\sigma \in P((x,y),i)}{c_{t,u,v}(\sigma)},
\ee
and
\be
\hat{P} =\sset{\sigma \in P((x,y),i): c_{t,u,v}(\sigma)}.
\ee
Then
\be
p_{(x,y),i} = \sum_{t=k}^{k+(s+1)m-1} \sum_{u=t}^{k+rm-1} \sum_{v=k}^{u} p_{t,u,v} p_s p_{i-1-s} p_g, \label{pixie}
\ee
where
\be
p_s = \probover{\sigma \in \hat{P}}{b_s |\sigma} \label{allsneighbors},
\ee
\be
p_{i-1-s} = \probover{\sigma \in \hat{P}}{\neg a_{s+1} \land \ldots \land \neg a_{i-1} |\sigma}, \hbox{ and}\label{allineighbors}
\ee
\be
p_g = \probover{\sigma \in \hat{P}}{\wrong{\gws{A_s \cup \sset{i}}} |\sigma}. \label{ptuv}
\ee
To see why, compare this to Equation \ref{triangle2}. For each $(t,u,v)$, we multiply the probability of $c_{t,u,v}$, which is $p_{t,u,v}$, by $p_s$, $p_{i-1-s}$, and $p_g$, each conditioned on $c_{t,u,v}$. (Taking probabilities over $\hat{P}$ conditions on $c_{t,u,v}$.) Together, the conditions in $p_s$, $p_{i-1-s}$, and $p_g$ are equivalent to the condition in Equation \ref{triangle2}, because $b_s \land \neg a_{s+1} \land \ldots \land \neg a_{i-1} |\sigma$ equals $b_{i-1}|\sigma$. The limits of summation for $t$ and $u$ follow from the fact that, with $(x,y) \in V_i|\sigma$, there are $(s+1)m-1$ remaining non-voter assignments and $rm-1$ remaining validation set assignments for each $\sigma$ in $P((x,y),i)$. 

To compute $p_{t,u,v}$, note that with $(x,y) \in V_i|\sigma$, for $\sigma \in P((x,y),i)$, there are $n-1$ remaining assignments, including $m-1$ to $V_i|\sigma$, $m$ for each other validation subset, and $n - rm$ for $F-V|\sigma$. This includes $n - (s+1)m$ voters and $(s+1)m-1$ non-voters. So
\be
p_{t,u,v} = 
\ee
\be
\frac{{{n-(s+1)m}\choose{k}} {{(s+1)m-1}\choose{t-k}}}{{{n-1}\choose{t}}} \left(\frac{k}{t}\right) \frac{{{n-(s+1)m-k}\choose{v-k}} {{(s+1)m-1-(t-k)}\choose{u-t-(v-k)}}}{{{n-1-t}\choose{u-t}}} \frac{{{n-rm}\choose{k}} {{(r-(s+1))m}\choose{v-k}}}{{{n - (s+1)m}\choose{v}}} 
\ee
\be
\sum_{z=0}^{\min(k,v-k)} \left(\frac{{{v-k}\choose{z}} {{k}\choose{k-z}}}{{{v}\choose{k}}} \frac{z}{u-t}\right). 
\ee
The terms are the probabilities of the following conditions, respectively, each conditioned on the previous terms' conditions:
\begin{enumerate}
\item There are exactly $k$ voters among the first $t$ neighbors. 
\item Neighbor $t$ is one of those $k$ voters. 
\item The first $u$ neighbors include exactly $v$ voters. 
\item Exactly $k$ of the $v$ voters are in $F-V|\sigma$. 
\item Neighbor $u$ is from $F-V|\sigma$. (The sum is over the number $z$ of neighbors $t+1$ to $u$ in $F-V|\sigma$.) If $\frac{z}{u-t} = \frac{0}{0}$, then treat it as one. 
\end{enumerate}

Now consider the three probabilities $p_s$, $p_{i-1-s}$, and $p_g$. The first is the probability that the validation subsets $V_1|\sigma, \ldots, V_s|\sigma$ are all represented among the nearer neighbors to $(x,y)$ than the $\kth$ nearest neighbor from $F-V|\sigma$. Since we condition on $c_{t,u,v}$ (by taking the probability only over $\sigma \in \hat{P}$, for which $c_{t,u,v}$ holds), the condition is that among the neighbors assigned $u-v$ of the $(s+1)m-1$ non-voter positions, each of $s$ sets of $m$ positions is represented. Use inclusion and exclusion, counting all ways to select the $u-v$ neighbors, subtracting ways to select the $u-v$ neighbors without drawing from each set $V_1|\sigma, \ldots, V_s|\sigma$, adding those that avoid drawing from each pair of sets, and so on:
\be
p_s = \sum_{j=0}^{s} (-1)^j {{s}\choose{j}} {{(s+1-j)m-1}\choose{u-v}} {{(s+1)m-1}\choose{u-v}}^{-1}.
\ee

Similarly, the condition for $p_{i-1-s}$, given $c_{t,u,v}$, is that all of $V_{s+1}|\sigma, \ldots, V_{i-1}|\sigma$ are represented among the $v-k$ voters with positions in $V_{s+1} \cup \ldots \cup V_{i-1} \cup V_{i+1} \cup \ldots \cup V_r|\sigma$. (The other $k$ voters are in $F-V|\sigma$.) Once again, use inclusion and exclusion:
\be
p_{i-1-s} = \sum_{j=0}^{i-1-s} (-1)^j {{i-1-s}\choose{j}} {{(r-s-1-j)m}\choose{v-k}} {{(r-s-1)m}\choose{v-k}}^{-1}.
\ee

The condition for $p_g$, given $c_{t,u,v}$, is that at least $\frac{k+1}{2}$ of the nearest $k$ voters, of which the last is neighbor $t$, have labels that disagree with $y$. Let $y_j$ be the label of neighbor $j$. Let $d_j$ count the labels among neighbors 1 to $j$ that disagree with $y$. Use $b$ to count how many neighbors with labels that disagree with $y$ are among the $k-1$ voters nearer to $(x,y)$ than neighbor $t$. Then
\be
p_g = \sum_{b=\frac{k+1}{2} - I(y_t\not=y)}^{k-1} {{d_{t-1}}\choose{b}} {{t-1-d_{t-1}}\choose{k-1-b}} {{t-1}\choose{k-1}}^{-1}.
\ee

Substitute Equation \ref{pixie} into Equation \ref{triangle3} to get an equation for $\phone$:
\be
\phone = \meanover{(x,y) \in F}{\sum_{i=1}^{r} \sum_{s=0}^{i-1}  {{i-1}\choose{s}} (-1)^s \sum_{t=k}^{k+(s+1)m-1} \sum_{u=t}^{k+rm-1} \sum_{v=k}^{u} p_{t,u,v} p_s p_{i-1-s} p_g}.
\ee
For $\tilde{p}'_I$, reduce the upper limits of summation for $t$ and $u$ to $k+r$:
\be
\tilde{p}'_I = \meanover{(x,y) \in F}{\sum_{i=1}^{r} \sum_{s=0}^{i-1}  {{i-1}\choose{s}} (-1)^s \sum_{t=k}^{k+r} \sum_{u=t}^{k+r} \sum_{v=k}^{u} p_{t,u,v} p_s p_{i-1-s} p_g}.
\ee
To compute this value, notice that only $p_g$ depends on values that are specific to each example $(x,y)$ -- the values $d_{t-1}$ and $I(y_t\not=y)$. Since $p_g$ only depends on those values and $t$, we can rearrange the sum:
\be
\tilde{p}'_I = \meanover{(x,y) \in F}{\sum_{t=k}^{k+r} p_g q(t)}, \label{pandq}
\ee
where
\be
q(t) = \sum_{i=1}^{r} \sum_{s=0}^{i-1}  {{i-1}\choose{s}} 2^{-(i-1)} (-1)^s  \sum_{u=t}^{k+r} \sum_{v=k}^{u} p_{t,u,v} p_s p_{i-1-s}.
\ee
To compute $q(t)$, first compute and store $p_{i-1-s}$ for all feasible $(i,s,v)$ and $p_s$ for all feasible $(s,u,v)$. Next, compute and store the last term of $p_{t,u,v}$ for all feasible $(v, u-t)$, then use those values to compute and store $p_{t,u,v}$ for all feasible $(s,t,u,v)$. This requires O($r^4$) computation and storage. Then compute $q(t)$ for each $t \in \sset{k, \ldots, k+r}$ by iterating through the sums and using the pre-computed values for $p_{t,u,v}$, $p_s$, and $p_{i-1-s}$. This requires O($r^4$) computation. 

To compute $\tilde{p}'_I$, first compute $p_g$ for all feasible $(t,d_{t-1},I(y_t\not=y))$. This requires O($rk(k+r)$) computation and O($r(k+r)$) storage. Then, for each $(x,y) \in F$, find its $k+r$ nearest neighbors in $F-V$, use the neighbors' labels to compute $d_{t-1}$ and $I(y_t\not=y)$ for $t \in \sset{k, \ldots, k+r}$. This requires O($k+r$) computation. Then compute the sum over $t$ in Equation \ref{pandq}, using $d_{t-1}$ and $I(y_t\not=y)$ values to select precomputed $p_g$ values and using the precomputed $q(t)$ values. This produces a sample value for $(x,y)$. Average those sample values over $(x,y) \in F$ to compute $\tilde{p}'_I$.

Using this method, aside from the time to find the $k+r$ nearest neighbors to each in-sample example, the time complexity is O($\max(r^4, rk(k+r), n (k+r))$) and the storage complexity is O($\max(r^4, r(k+r))$). If $k \in \hbox{O}(\ln n)$ and $r \in \hbox{O}(\ln n)$ and $n > (\ln n)^3$, then this is O($n \ln n$) time and O($(\ln n)^4$) storage.

\bibliographystyle{elsarticle-num} 
\bibliography{bax}

\begin{thebibliography}{10}
\expandafter\ifx\csname url\endcsname\relax
  \def\url#1{\texttt{#1}}\fi
\expandafter\ifx\csname urlprefix\endcsname\relax\def\urlprefix{URL }\fi
\expandafter\ifx\csname href\endcsname\relax
  \def\href#1#2{#2} \def\path#1{#1}\fi

\bibitem{vapnik71}
V.~Vapnik, A.~Chervonenkis, On the uniform convergence of relative frequencies
  of events to their probabilities, Theory of Probability and its Applications
  16 (1971) 264--280.

\bibitem{valiant84}
L.~G. Valiant, A theory of the learnable, Commun. ACM 27~(11) (1984)
  1134--1142.
\newblock \href {http://dx.doi.org/http://doi.acm.org/10.1145/1968.1972}
  {\path{doi:http://doi.acm.org/10.1145/1968.1972}}.

\bibitem{littlestone86}
N.~Littlestone, M.~Warmuth, Relating data compression and learnability,
  unpublished manuscript, University of California Santa Cruz (1986).

\bibitem{floyd95}
S.~Floyd, M.~Warmuth, Sample compression, learnability, and the
  {V}apnik-{C}hervonenkis dimension, Machine Learning 21~(3) (1995) 1--36.

\bibitem{blum03}
A.~Blum, J.~Langford, {P}{A}{C}-{M}{D}{L} bounds, in: Proceedings of the 16th
  Annual Conference on Computational Learning Theory (COLT), 2003, pp.
  344--357.

\bibitem{bax_compression}
E.~Bax, Nearly uniform validation improves compression-based error bounds,
  Journal of Machine Learning Research 9 (2008) 1741--1755.

\bibitem{baxcallejas}
E.~Bax, A.~Callejas, An error bound based on a worst likely assignment, Journal
  of Machine Learning Research 9 (2008) 581--613.

\bibitem{langford05}
J.~Langford, Tutorial on practical prediction theory for classification,
  Journal of Machine Learning Research 6 (2005) 273--306.

\bibitem{cover67}
T.~M. Cover, P.~E. Hart, Nearest neighbor pattern classification, IEEE
  Transactions on Information Theory 13 (1967) 21--27.

\bibitem{cover68}
T.~M. Cover, Rates of convergence for nearest neighbors procedures, in: B.~K.
  Kinariwala, F.~F. Kuo (Eds.), Proceedings of the Hawaii International
  Conference on System Sciences, University of Hawaii Press, 1968, pp.
  413--415.

\bibitem{psaltis94}
D.~Psaltis, R.~Snapp, S.~Venkatesh, On the finite sample performance of the
  nearest neighbor classifier, IEEE Transactions on Information Theory 40~(3)
  (1994) 264--280.

\bibitem{devroye96}
L.~Devroye, L.~Gy{\"o}rfi, G.~Lugosi, A Probabilistic Theory of Pattern
  Recognition, Springer, 1996.

\bibitem{duda01}
R.~O. Duda, P.~E. Hart, D.~G. Stork, Pattern Classification, Wiley, 2001.

\bibitem{hastie09}
T.~Hastie, R.~Tibshirani, J.~Friedman, The Elements of Statistical Learning:
  Data Mining, Inference, and Prediction, Second Edition, Springer, 2009.

\bibitem{rogers78}
W.~Rogers, T.~Wagner, A finite-sample distribution-free performance bound for
  local discrimination rules, Annals of Statistics 6 (1978) 506--514.

\bibitem{devroye79}
L.~Devroye, T.~Wagner, Distribution-free inequalities for the deleted and
  holdout estimates, IEEE Transactions on Information Theory 25 (1979)
  202--207.

\bibitem{bax15}
E.~Bax, Y.~Le, \href{http://arxiv.org/abs/1510.02676}{Some theory for practical
  classifier validation}, Baylearn.
\newline\urlprefix\url{http://arxiv.org/abs/1510.02676}

\bibitem{hoeffding63}
W.~Hoeffding, Probability inequalities for sums of bounded random variables,
  Journal of the American Statistical Association 58~(301) (1963) 13--30.

\bibitem{boucheron13}
S.~Boucheron, G.~Lugosi, P.~Massart, Concentration Inequalities -- A
  Nonasymptotic Theory of Independence, Oxford University Press, 2013.

\bibitem{bax12}
E.~Bax, Validation of $k$-nearest neighbor classifiers, IEEE Transactions on
  Information Theory 58~(5) (2012) 3225--3234.

\bibitem{chvatal79}
V.~Chv\'atal, The tail of the hypergeometric distribution, Discrete Mathematics
  25~(3) (1979) 285--287.

\bibitem{skala13}
M.~Skala, \href{https://arxiv.org/abs/1311.5939v1}{Hypergeometric tail
  inequalities: ending the insanity}, arXiv arXiv:1311.5939v1.
\newline\urlprefix\url{https://arxiv.org/abs/1311.5939v1}

\bibitem{bax_average}
E.~Bax, F.~Kooti, \href{https://arxiv.org/pdf/1610.01234v1.pdf}{Ensemble
  validation: Selectivity has a price, but variety is free}, Baylearn.
\newline\urlprefix\url{https://arxiv.org/pdf/1610.01234v1.pdf}

\bibitem{maurer09}
A.~Maurer, M.~Pontil, Empirical bernstein bounds and sample-variance
  penalization, 22nd Annual Conference on Learning Theory (COLT).

\bibitem{bennett62}
G.~Bennett, Probability inequalities for the sum of independent random
  variables, Journal of the American Statistical Association 57~(297) (1962)
  33--45.

\bibitem{bax98}
E.~Bax, Validation of average error rate over classifiers, Pattern Recognition
  Letters (1998) 127--132.

\bibitem{mcallester99}
D.~A. McAllester, Pac-bayesian model averaging, in: In Proceedings of the
  Twelfth Annual Conference on Computational Learning Theory, ACM Press, 1999,
  pp. 164--170.

\bibitem{langford01}
J.~Langford, M.~Seeger, N.~Megiddo, An improved predictive accuracy bound for
  averaging classifiers, in: In Proceeding of the Eighteenth International
  Conference on Machine Learning, 2001, pp. 290--297.

\bibitem{vapnik98}
V.~Vapnik, Statistical Learning Theory, John Wiley \& Sons, 1998.

\bibitem{kedem12}
D.~Kedem, S.~Tyree, F.~Sha, G.~R. Lanckriet, K.~Q. Weinberger,
  \href{http://papers.nips.cc/paper/4840-non-linear-metric-learning.pdf}{Non-linear
  metric learning}, in: F.~Pereira, C.~J.~C. Burges, L.~Bottou, K.~Q.
  Weinberger (Eds.), Advances in Neural Information Processing Systems 25,
  Curran Associates, Inc., 2012, pp. 2573--2581.
\newline\urlprefix\url{http://papers.nips.cc/paper/4840-non-linear-metric-learning.pdf}

\bibitem{sen2008}
P.~Sen, G.~Namata, M.~Bilgic, L.~Getoor, B.~Gallagher, T.~Eliassi-Rad,
  Collective classification in network data, AI Magazine 29~(3) (2008) 93--106.

\bibitem{macskassy2007}
S.~A. Macskassy, F.~Provost, Classification in networked data: A toolkit and a
  univariate case study, Journal of Machine Learning Research 8 (2007)
  935--983.

\bibitem{london12}
B.~London, B.~Huang, L.~Getoor, Improved generalization bounds for large-scale
  structured prediction, in: NIPS Workshop on Algorithmic and Statistical
  Approaches for Large Social Networks, 2012.

\bibitem{li12}
J.~Li, A.~Sonmez, Z.~Cataltepe, E.~Bax, Validation of network classifiers,
  Structural, Syntactic, and Statistical Pattern Recognition Lecture Notes in
  Computer Science 7626 (2012) 448--457.

\bibitem{bax13}
E.~Bax, J.~Li, A.~Sonmez, Z.~Cataltepe, Validating collective classification
  using cohorts, NIPS Workshop on Frontiers of Network Analysis: Methods,
  Models, and Applications.

\bibitem{cristianini00}
N.~Cristianini, J.~Shawe-Taylor, An Introduction to Support Vector Machines and
  Other Kernel-Based Learning Methods, Cambridge University Press, 2000.

\bibitem{joachims02}
T.~Joachims, Learning to Classify Text using Support Vector Machines, Kluwer
  Academic Publishers, 2002.

\bibitem{marchand01}
M.~Marchand, J.~Shawe-Taylor, Learning with the set covering machine, in:
  Proceedings of the Eighteenth International Conference on Machine Learning
  (ICML 2001), 2001, pp. 345--352.

\bibitem{mullin00}
M.~Mullin, R.~Sukthankar, Complete cross-validation for nearest neighbor
  classifiers, Proceedings of the Seventeenth International Conference on
  Machine Learning (2000) 639--646.

\end{thebibliography}

\end{document}